\documentclass{article}
\pdfoutput=1

\PassOptionsToPackage{numbers, compress}{natbib}
\usepackage[final]{neurips_2023}
\usepackage{graphicx}
\graphicspath{ {./figures/} }
\usepackage{subcaption}
\usepackage{amsmath}
\usepackage{newtxmath}
\DeclareMathAlphabet{\mathpzc}{T1}{pzc}{m}{it}
\usepackage{algorithm}
\usepackage{algpseudocode}
\usepackage{booktabs}       
\usepackage{multirow}
\usepackage{enumitem}

\newcommand{\xhdr}[1]{\vspace{0em}\noindent{{\bf #1.}}}
\newcommand{\ie}{\textit{i.e., }}
\newcommand{\eg}{\textit{e.g., }}

\usepackage{colortbl}

\definecolor{Gray}{gray}{0.9}
\definecolor{LightCyan}{rgb}{0.88,1,1}
\newcolumntype{a}{>{\columncolor{Gray}}c}
\newcolumntype{b}{>{\columncolor{white}}c}

\usepackage{amsfonts}
\usepackage{mathtools}
\usepackage{enumitem}
\usepackage{comment}
\usepackage{mdframed}
\usepackage{wrapfig}
\usepackage{ntheorem}
\theoremseparator{}
\theoremstyle{nonumberplain}
\usepackage{ntheorem}
\theoremseparator{}
\theoremstyle{nonumberplain}

\usepackage{tcolorbox}

\tcbuselibrary{skins} 

\newtcolorbox{dialogbox}{
    enhanced,
    boxrule=1pt, 
    colback=black!10, 
    colframe=black, 
    left=1pt, 
    right=1pt, 
    top=3pt, 
    bottom=3pt, 
}

\usepackage[utf8]{inputenc} 
\usepackage[T1]{fontenc}    
\usepackage{hyperref}       
\usepackage{url}            
\usepackage{booktabs}       
\usepackage{amsfonts}       
\usepackage{nicefrac}       
\usepackage{microtype}      
\usepackage{xcolor}         

\title{Quantifying Uncertainty in Natural Language Explanations of Large Language Models}

\author{%
  Sree Harsha Tanneru \\
  \texttt{sreeharshatanneru@g.harvard.edu} \\
  \And
  Chirag Agarwal \\
  \texttt{chiragagarwall12@gmail.com} \\
  \And
  Himabindu Lakkaraju \\
  \texttt{hlakkaraju@seas.harvard.edu} \\
  \AND
  \vspace{-0.25in}  \\
  Harvard University
}

\begin{document}

\maketitle

\begin{abstract}
\vspace{-0.01in}
\looseness=-1 Large Language Models (LLMs) are increasingly used as powerful tools for several high-stakes natural language processing (NLP) applications. Recent prompting works claim to elicit intermediate reasoning steps and key tokens that serve as proxy explanations for LLM predictions. However, there is no certainty whether these explanations are reliable and reflect the LLM’s behavior. In this work, we make one of the first attempts at quantifying the uncertainty in explanations of LLMs. To this end, we propose two novel metrics --- \textit{Verbalized Uncertainty} and \textit{Probing Uncertainty} --- to quantify the uncertainty of generated explanations. While verbalized uncertainty involves prompting the LLM to express its confidence in its explanations, probing uncertainty leverages sample and model perturbations as a means to quantify the uncertainty. Our empirical analysis of benchmark datasets reveals that verbalized uncertainty is not a reliable estimate of explanation confidence. Further, we show that the probing uncertainty estimates are correlated with the faithfulness of an explanation, with lower uncertainty corresponding to explanations with higher faithfulness. Our study provides insights into the challenges and opportunities of quantifying uncertainty in LLM explanations, contributing to the broader discussion of the trustworthiness of foundation models.
\end{abstract}

\section{Introduction}
\looseness=-1 Large Language Models (LLMs), such as GPT4~\citep{openai2023gpt}, Bard~\citep{manyika2023overview}, Llama-2~\citep{LLAMA2}, and Claude-2~\citep{anthropic2023}, have garnered significant attention and are employed across a wide range of applications, including chat-bots, computational biology, creative work, and law~\citep{kaddour2023challenges} due to their impressive natural language understanding and generation capabilities. However, state-of-the-art LLMs are complex models with billions of parameters, where their inner working mechanisms are not fully understood yet, making them less trustworthy amongst relevant stakeholders. This lack of transparency causes hindrance to deploying LLMs in high-stakes decision-making applications, where the consequences of incorrect decisions are severe and could result in the generation of harmful content, misdiagnosis~\citep{zhang2023potential}, and hallucinations~\citep{Ji_2023,weidinger2021ethical}. The lack of user trust demands the development of robust explanation techniques to gain insights into how these powerful LLMs work.

\looseness=-1 Previous works for explaining language models can be broadly categorized into perturbation-based methods~\citep{li2016visualizing,DBLP:journals/corr/LiMJ16a}, gradient-based methods~\citep{kindermans2017unreliability,pmlr-v70-sundararajan17a}), attention-based methods~\citep{derose2020attention,vig2019visualizing}, example-based methods~\citep{jin2020bert,treviso-etal-2023-crest,wang2022semattack, wu-etal-2021-polyjuice}, and Natural Language Explanations (NLEs)~\citep{wei2023chainofthought}. While most of the above methods require white-box access to models (\eg model gradients and prediction logits), NLEs generated by LLMs enable us to understand the behavior of these models even when the models are closed-source. For instance, Chain-of-Thought (CoT)~\citep{wei2023chainofthought} explanations, a popular class of NLEs generated by LLMs show the step-by-step reasoning process leading to the outputs generated by these models. While CoTs and other natural language explanations generated by LLMs often seem quite plausible and believable~\cite{turpin2023language}, recent works have demonstrated that these natural language explanations may not always faithfully capture the underlying behavior of these models~\cite{turpin2023language}. However, there is little to no work that focuses on deciphering if and to what extent the generated NLEs faithfully capture the behavior of the underlying model. One way to address this problem is to quantify the uncertainty in the NLEs generated by LLMs. However, this critical direction remains unexplored.
\looseness=-1 
Prior works on uncertainty estimation in the context of LLMs have only focused on providing uncertainty estimates (\ie confidence) corresponding to the responses (answers) generated by LLMs~\citep{xiong2023llms}. While uncertainty in LLM predictions has been studied using external calibrators~\citep{jiang-etal-2021-know}, model fine-tuning~\citep{lin2022teaching}, and non-logit-based approaches~\citep{xiong2023llms}, there is little to no work on estimating the uncertainty of LLM explanations. Understanding the uncertainty in natural language explanations generated by LLMs is paramount to ensuring that these explanations are trustworthy and are not just plausible hallucinations. 

\xhdr{Present work} In this work, we make one of the attempts at quantifying the uncertainty in natural language explanations generated by LLMs. In particular, we propose two novel approaches -- \textit{Verbalized uncertainty} and \textit{Probing uncertainty} metrics -- to quantify the confidence of NLEs generated by large language models and compare their reliability. While verbalized uncertainty metrics focus on prompting a language model to express its uncertainty in the generated explanations, probing uncertainty metrics leverage different kinds of input perturbations (e.g., replacing words with synonyms, paraphrasing inputs) and measure the consistency of the resulting explanations. 
Using our proposed metrics, we provide the first definition of uncertainty estimation of language model explanations. In addition, our work also demonstrates key connections between \textit{uncertainty} and \textit{faithfulness} of natural language explanations generated by LLMs. 

\looseness=-1 We evaluate the effectiveness of our proposed metrics on three math word problems and two commonsense reasoning benchmark datasets and conduct experiments using different \textsc{Gpt} variants.
Our empirical results across these datasets and LLMs reveal the following key findings. 1) Verbalized uncertainty is not a reliable estimate of explanation confidence and LLMs often exhibit very high verbalized confidence in the explanations they generate. 2) Probing uncertainty is correlated with the predictive performance of the LLM, where correct answers from a model tend to generate more confident/less uncertain explanations. 3) A clear connection exists between the uncertainty and faithfulness of an explanation, where less uncertain explanations tend to be more faithful to the model predictions.

\begin{figure*}[h]
\begin{dialogbox}
Read the question, and assign each word an importance score between 0 and 100 of how important it is for your answer. The output format is as follows:\vspace{0.05in}

\textbf{Word}: [Word 1 here], \textbf{Importance}: [Your importance score here]\\
\dots \\
\textbf{Word}: [Word N here], \textbf{Importance}: [Your importance score here]\vspace{0.03in}

\looseness=-1\textcolor{purple}{\textbf{Final answer and overall confidence (0-100):}} [Your answer as a number here], [Your confidence here] \vspace{0.03in}

Note: The importance scores of all words should add up to 100. The overall confidence score indicates the degree of certainty you have about your importance scores. For instance, if your confidence level is 80\%, it means you are 80\% certain that importance scores assigned are correct.
Provide the answer in aforementioned format, and nothing else.\vspace{0.04in}

\textbf{Q:}~Jake has 11 fewer peaches than Steven. If Jake has 17 peaches. How many peaches does Steven have?\\

\textcolor{black!40!green}{\textbf{Answer:}}\\
\textbf{Word:} Jake, \textbf{Importance:} 20\%\\
\textbf{Word:} Steven, \textbf{Importance:} 20\%\\
\textbf{Word:} peaches, \textbf{Importance:} 60\%\\
\textcolor{purple}{\textbf{Final answer and overall confidence (0-100):}} 28, 100\%
    \end{dialogbox}
    \caption{\small\looseness=-1\xhdr{Template for generating token importance and its confidence} The prompt $Q_e$ appended to the original question $Q$ to elicit a token importance explanation $TI$. We ask the underlying LLM to verbally assign an importance score to each word in the question $Q$ and then provide the final answer $A$ with overall confidence.
    }\label{fig:vanillafeatureimportanceexplanationuncertainty}
\end{figure*}

\section{Preliminaries}
\label{sec:prelims}
\xhdr{Notations} 
Large language models typically have a single vocabulary $\mathcal{V}$ that represents a set of unique ``tokens'' (words or sub-words). Let $\mathcal{M}: Q \rightarrow A$ denote a language model mapping a sequence of $n$ question tokens $Q = (q_1, q_2, \dots, q_n)$ to sequence of $m$ answer tokens $A = (a_1, a_2, \dots, a_m)$, where $q_i$ and $a_i$ are text tokens belonging to the model vocabulary $\mathcal{V}$. In addition to the original question $Q$, we design specific prompts $Q_e$ to generate natural language explanation (NLE) $A_e$ from the language model $\mathcal{M}$.

\looseness=-1\xhdr{Uncertainty} Black-box LLMs do not provide access to parameter gradients or model logits, rendering traditional explainability techniques ineffective. To this end, most language models leverage NLEs, which are explanations generated from the language model to serve as proxy explanations and are a viable alternative. While NLEs are essentially a sequence of tokens sampled from the model that serve as explanations, there is an associated uncertainty for the generated explanations. Quantifying the uncertainty of these explanations is essential to estimate the reliability of generated NLEs. For the rest of the paper, we will use the term ``confidence score'' to refer to the uncertainty of an explanation, as determined by the language model.

\looseness=-1 \xhdr{Explanation Methods} We confine our study to two explanation methods --- Token Importance and Chain of Thought (CoT) explanations. While token importance explanations~\cite{li2016visualizing,wu-etal-2020-perturbed} aims to identify input tokens (refer to tokens $t$ in an input text $T$ for LLMs) that most contribute to a model's predictions, CoT explanations~\citep{wei2023chainofthought} focus on revealing the sequence of operations or reasoning steps $S_{i} \in S$ the language model $\mathcal{M}$ takes when processing the question $Q$ and arriving at its predictions, where $n_s = |S|$ denotes the total number of steps in a CoT explanation. For token importance explanation, we concatenate a prompt $Q_e$ to the given question  $Q$ using the template: ``\textit{Read the question and output the words important for your final answer\dots}''. Whereas, the prompt $Q_e$ to generate CoT explanations uses the following template: ``\textit{Read the question, give your answer by analyzing step by step, \dots}''. Please refer to Figures~\ref{fig:featureimportanceexplanation}-\ref{fig:cotexplanation} in appendix for more details.

\looseness=-1 We generate an answer from the LLM $\mathcal{M}$ as follows: $\mathcal{M}(Q) = A$. We also generate an explanation $A_e$ along with answer $A$ using the aforementioned template question $Q_e$ as: $\mathcal{M}(Q_e + Q) = A + A_e$.

\begin{figure*}[h]
\begin{dialogbox}
Read the question, give your answer by analyzing step by step, and assign a confidence level to each step and the final answer. The output format is as follows:\\
\textbf{Step 1}: [Your reasoning here], \textbf{Confidence}: [Your confidence here]\% \\
\textbf{Step 2}: [Your reasoning here], \textbf{Confidence}: [Your confidence here]\% \\
...\\
\textbf{Step N}: [Your reasoning here], \textbf{Confidence}: [Your confidence here]\% \vspace{0.03in}

\looseness=-1\textcolor{purple}{\textbf{Final answer and overall confidence (0-100):}}~[Your answer as a number here],~[Your confidence here]\%\vspace{0.03in}

Note: The confidence indicates the degree of certainty you have about your reasoning. For instance, if your confidence level is 80\%, it means you are 80\% certain that your reasoning is correct.
Provide the answer in aforementioned format, and nothing else.\vspace{0.04in}

\textbf{Q:}~Jake has 11 fewer peaches than Steven. If Jake has 17 peaches. How many peaches does Steven have?\\

\textcolor{black!40!green}{\textbf{Answer:}}\\
Step 1: Jake has 11 fewer peaches than Steven.~~Confidence: 100\%\\
Step 2: Jake has 17 peaches.~~Confidence: 100\%\\
Step 3: If Jake has 17 peaches, then Steven has 17 + 11 = 28 peaches.~~Confidence: 100\%\\
\textcolor{purple}{\textbf{Final answer and overall confidence (0-100):}} 28, 100\%
\end{dialogbox}
    \caption{\small\xhdr{Template for generating CoT explanation and its step-wise confidence} The prompt $Q_e$ appended to the original question $Q$ to elicit a CoT explanation. We ask the underlying LLM to verbally assign an importance score to each step of the CoT explanation and then provide the final answer $A$ with overall confidence.
    }\label{fig:vanillacotexplanationuncertainty}
\end{figure*}

\section{Quantifying Uncertainty in Explanations}
\label{sec:method}
\looseness=-1 Next, we describe our metrics which aim to estimate the uncertainty in token importance and CoT explanations generated by LLMs.

\looseness=-1\xhdr{Problem formulation (Uncertainty in  Explanations)}~\textit{Given a question-answer pair and prompt $Q_e$ to generate natural language explanation $A_e$ from the model $\mathcal{M}: (Q, Q_e) \rightarrow (A, A_e)$, we aim to develop an uncertainty function $\textsc{Unc}: A_e \rightarrow [0, 1]$, which maps a generated explanation $A_e$ to a scalar score that determines the uncertainty in the generated explanation, \ie
}
$$\text{Uncertainty} = \textsc{Unc}(A_e),$$
where $\mathcal{M}(Q_e + Q) = A + A_e $.

As mentioned before, we confine our study to two natural language explanation methods -- Token Importance and CoT. We use $\text{TI}_q: \{w~|~w \in Q\}$ to denote a token importance explanation which is a subset of words in the question $Q$ that are important for predicting the answer $A$ and $\text{CoT}_q: \{(S_1, c_1) \rightarrow (S_2, c_2) \dots \rightarrow (S_{n_s}, c_{n_s})\}$ a CoT explanation for a prediction $A$ from question $Q$. Here $S_i = (s_1, s_2, \dots s_{n_s})$ is a text sequence denoting the natural language reasoning and $c_i \in [0, 1]$ is the LLM's confidence of CoT step $S_i$.


\subsection{Verbalized Uncertainty}
\label{sec:verbalized}
A straightforward approach to elicit uncertainty of an explanation is to directly request the LLM $\mathcal{M}$ to output a confidence score for the explanation ranging from 0\% to 100\%. By directly soliciting the model's self-assessment of uncertainty, this approach seeks to extract explicit uncertainty information inherent in the model. We provide the template of the prompts for confidence elicitation for token importance and CoT explanations in Figures~\ref{fig:vanillafeatureimportanceexplanationuncertainty}-\ref{fig:vanillacotexplanationuncertainty}. For token importance, we ask the underlying LLM to verbally assign an importance score to each word in the question $Q$ and then provide the final answer of the question with overall confidence (see Fig.~\ref{fig:vanillafeatureimportanceexplanationuncertainty}). In contrast, for CoT explanations (see Fig.~\ref{fig:vanillacotexplanationuncertainty}), we ask the LLM to assign verbalized confidence to each step in the CoT reasoning and the final answer. 

\subsection{Probing Uncertainty}
\label{sec:probing}
Verbalized uncertainty elicits confidence in an explanation by directly requesting the underlying LLM to output a confidence score in a given range. In contrast, for estimating uncertainty using probing, we leverage the consistency of explanations as a measure to estimate the uncertainty in explanations generated by a language model $\mathcal{M}$. More specifically, let $A_e$ denote the natural language explanation generated by the model $\mathcal{M}$ for a given question $Q$ and $[A_{e_1}, A_{e_2}, \dots, A_{e_N}]$ be $N$ explanations generated for $N$ perturbation of the same question using its local neighborhood. Next, we describe two different perturbation strategies to generate $N$ explanations for a given question and answer.

\xhdr{Sample Probing} Motivated by the local neighborhood approximation works in XAI~\citep{ribeiro2016should,smilkov2017smoothgrad}, we propose uncertainty metrics that leverage the consistency of a model in generating the explanation in a local neighborhood. Here, we presume that the local behavior of the underlying LLM is consistent for perturbed samples of the original question and gradually introduce perturbations in the questions by \textit{paraphrasing} the original question $Q$. Given a question $Q$, we paraphrase the question into $N$ different forms $\{Q_1, Q_2, \dots, Q_N\}$, such that each paraphrased question $Q_i$ is semantically equivalent to Q, and the true reasoning process remains the same, \ie given a question: ``\textit{Jake has 11 fewer peaches than Steven. If Jake has 17 peaches. How many peaches does Steven have?}'', some of its local paraphrased counterparts used to calculate uncertainty in explanations are i)\textit{...What is the number of peaches Steven has?} ii)\textit{...How many peaches is Steven in possession of?} iii)\textit{...How many peaches does Steven possess?} Next, we generate the explanations using the LLM by probing the model using the paraphrased questions $Q_i$. Mathematically,
\begin{equation}
\mathcal{M}(Q_e + Q_i) = A_i + A_{e_i} \; ; \; i = 1, 2, \dots, N
\end{equation}
where $Q_i$ is a paraphrased form of question $Q$, $Q_e$ is the prompt to generate explanations, and $A_{e_i}$ is the corresponding generated explanation.

\xhdr{Model Probing} In contrast to sample uncertainty, where we quantify the uncertainty in explanations using the variance in the input questions,  model uncertainty addresses the uncertainty of LLM explanations due to the inherent stochasticity of the underlying language model $\mathcal{M}$. 
\begin{figure}[h]
    \centering
    \includegraphics[width=0.63\textwidth]{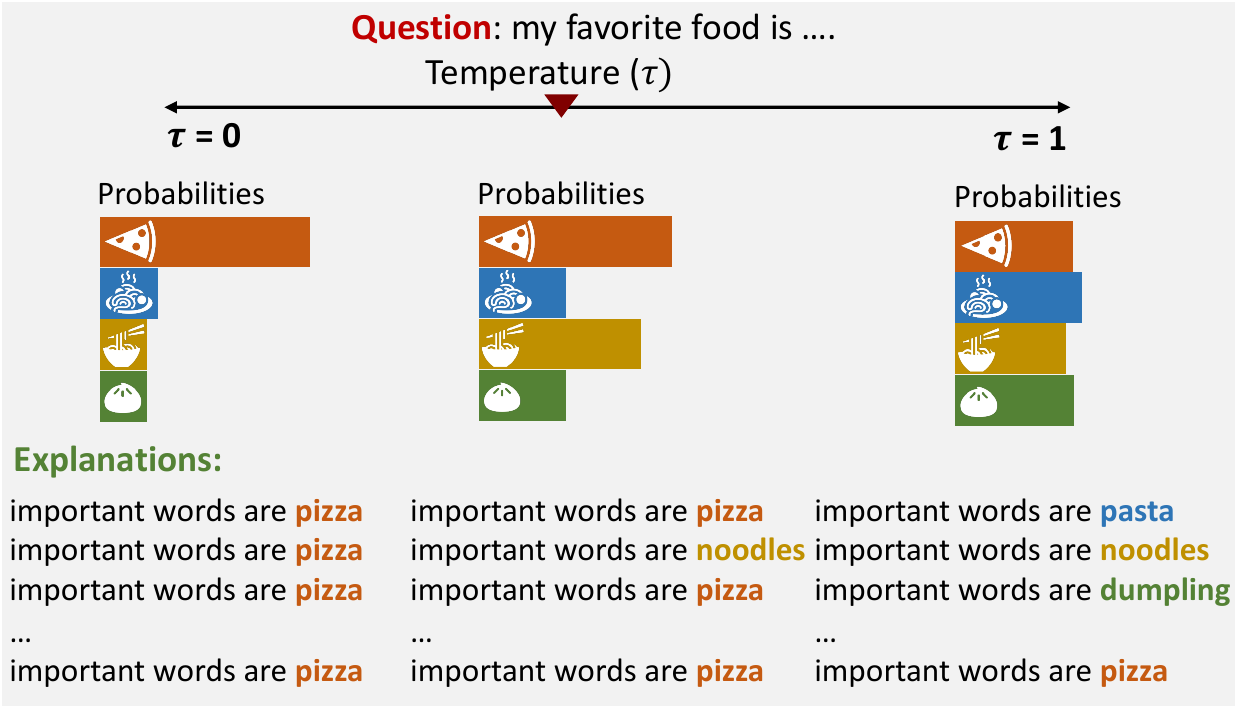}
    \caption{\small
    The impact of the temperature $\tau$ on model stochasticity. We find that as $\tau$ increases, the stochasticity in model responses increases. $\tau{=}0$ gives near-deterministic answers to a question, whereas $\tau{=}1$ gives a distribution of answers. 
    }
    \label{fig:temp}
\end{figure}
More specifically, we use the ``temperature'' parameter $\tau$ present in most LLMs that control the randomness in the generated answers by using the probability distribution of each generated token. A high value of temperature indicates an even distribution among all tokens, and a lower value of temperature indicates a sharper distribution (see Figure~\ref{fig:temp}). As the temperature parameter increases, the language model becomes more creative and stochastic in the generated explanations. Intuitively, the temperature parameter affects the sampling process when generating answers from the model. For a given question $Q$ (say ``\textit{my favorite food is \dots}''), we sample $N$ answers and their corresponding explanations, $\{A_i, A_{e_i}\} \forall i \in {1, 2, \dots, N}$ from the language model $\mathcal{M}$. Mathematically, we can denote this using:
\begin{equation}    
    \mathcal{M}(Q_e + Q) = A_i + A_{e_i} \; ; \; i \in \{1, 2, \dots, N\}
\end{equation}
where $A_i$ is the $i^{\text{th}}$ answer generated by the LLM for a given temperature $\tau$ and $A_{e_i}$ is its respective explanation.
\subsubsection{Token Importance Uncertainty}
\label{sec:estimation}
Using the above sample and model perturbation strategies, $N$ perturbed natural language explanations $A_{e_i}$ are generated for a given question $Q$, answer $A$, original explanation $A_{e}$. Next, we describe the metrics for estimating explanation confidence from these perturbed explanations.

\looseness=-1 We define the uncertainty in token importance explanations as the mean agreement between perturbed explanations and the original explanation. Two token importance explanations are said to agree with each other if they employ the same set of important words to arrive at a prediction. To quantify token importance uncertainty, we use token agreement and token rank metrics. While token agreement computes the fraction of important tokens that are common between two different explanations, token rank measures the fraction of important tokens that have the same position in their respective rank orders. The token rank (TR) metric is defined below:
\begin{equation}
    \begin{aligned}
        \text{TR}(\text{TI}_{i}, \text{TI}_{j}, k) &{=} \frac{1}{k} \Big(\bigcup_{s \in S} \{s | s \in \text{Tokens}(\text{TI}_{i}, k) \land s \in \\
        & \text{Tokens}(\text{TI}_{j}, k) \land \text{R}(\text{TI}_{i}, s) {=} \text{R}(\text{TI}_{j}, s) \}\Big),
    \end{aligned}
    \label{equation:tokenrankagreement}
\end{equation}
\looseness=-1 where $\text{TI}_{i}$ and $\text{TI}_{j}$ are any two given token important explanations, $\text{Tokens}(\text{TI}_{i}, k)$ is the first $k$ tokens in explanation $\text{TI}_i$, $k$ denotes the topK tokens a user wants as explanations, and $\text{R}(\cdot)$ function gives the rank of the word $s$ in a token importance explanation $\text{TI}$. The uncertainty in token importance explanation is defined as the mean agreement between the perturbed explanations $\text{TI}_{e_i}$ and the original explanation $\text{TI}_{\text{original}}$.
\begin{equation}
        \textsc{Unc}_{\text{TI}} = \dfrac{1}{N} \sum_{i=1}^{N} \text{TR}(\text{TI}_{e_i}, \text{TI}_{\text{original}}, k),
\end{equation}


\subsubsection{Chain of Thought Uncertainty}
While the agreement between token importance explanations is intuitive, the agreement between the chain of thought explanations is non-trivial as each explanation has a sequence of steps $S_i$ in natural language as output explanations. To check if the two steps in CoT explanations are equivalent, we propose using pre-trained sentence encoder models~\citep{reimers2019sentencebert}. Let us consider two CoT explanations that generate $N_a$ and $N_b$ steps in their respective explanations, \ie $(\text{CoT}_{a} = (s_{a_1}, s_{a_2}, \dots, s_{a_{N_a}})$ and $\text{CoT}_{b} = (s_{b_1}, s_{b_2}, \dots, s_{b_{N_b}})$. We define CoT agreement metric (CoTA) that measures the agreement between any two given CoT explanations as:

\begin{equation}
\begin{aligned}
    \text{CoTA}(\text{CoT}_{a}, \text{CoT}_{b}) &= \dfrac{1}{N_a + N_b} \Big(\sum_{i=1}^{N_a} \max_{j \in {1, \dots, N_{b}}} \text{E}(s_{a_i}, s_{b_j}) \\
    & + \sum_{j=1}^{N_b} \max_{i \in {1, \dots, N_{a}}} \text{E}(s_{a_i}, s_{b_j})\Big),
\end{aligned}
\label{eq:cota}
\end{equation}

The intuition behind the above metric is that for every step in the a CoT explanation, we check if there exists a step in other CoT explanation which agrees with it. $\text{E}(\cdot, \cdot)$ denotes the entailment model that focuses on the task of textual entailment or natural language inference (NLI). The goal of NLI is to determine the logical relationship between two sentences, usually framed as ``entailment'', ``contradiction'', or ``neutral''.  Formally, the entailment score between two explanation steps is defined as:
$$\text{E}(s_i, s_j) =
\begin{cases}
  1 & \text{if statements entail each other} \\
  0 & \text{if statements do not entail each other}
\end{cases}$$

Finally, the uncertainty in the CoT explanation is calculated as the mean agreement of the perturbed chain of thought explanations with the original explanation.
\begin{equation}
    \begin{aligned}
        \textsc{Unc}_{\text{CoT}} &= \dfrac{1}{N} \sum_{i=1}^{N} \text{CoTA}(\text{CoT}_{i}, \text{CoT}_{\text{original}})
    \end{aligned}
    \label{eq:cotuncertainty}
\end{equation}
To summarize, we introduce a metric for calculating the agreement between two CoT explanations (Eq.~\ref{eq:cota}). In addition, we generate $N$ perturbed explanations for a question, and calculate the mean agreement of perturbed explanations with the original explanation to estimate explanation uncertainty (Eq.~\ref{eq:cotuncertainty}).




\section{Experiments}
\label{sec:expt}
Next, we validate the effectiveness of our proposed uncertainty metric which amounts to asking: \textit{What is the uncertainty in explanations generated by state-of-the-art LLMs with respect to different explanation methods?} More specifically, we focus on the following research questions: RQ1) Does verbalized uncertainty estimation depict overconfidence in LLMs? RQ2) Is there a relation between uncertainty and faithfulness of an explanation? RQ3) How does explanation confidence vary for correct and incorrect answers? RQ4) Are changes in the metric parameters necessary for quantifying uncertainty in explanations?

\subsection{Datasets and Experimental Setup}
We first describe the datasets and large language models used to study the uncertainty in explanations and then outline the experimental setup.

\looseness=-1\xhdr{Datasets} We conduct experiments using three math word problem and two commonsense reasoning benchmark datasets. i) the \textbf{GSM8K} dataset that comprises several math word problems~\citep{cobbe2021training}, ii) the \textbf{SVAMP} dataset contains math word problems with varying structures~\citep{patel2021nlp}, iii) the \textbf{ASDiv} dataset consisting of diverse math word problems~\citep{miao2021diverse}, iv) the \textbf{StrategyQA}~\citep{geva2021did} requires a language model to deduce a multi-step reasoning strategy to answer questions and v) the \textbf{Sports Understanding} dataset, which is a specialized evaluation set from the BIG-bench~\citep{srivastava2022beyond} that involves determining whether a sentence relating to sports is plausible or implausible.


\xhdr{Large language models} We generate and evaluate the uncertainty in explanations by generating explanations using three large language models --- Instruct\textsc{Gpt}, \textsc{Gpt}-3.5, and \textsc{Gpt}-4.

\xhdr{Performance metrics} Some recent works ~\citep{atanasova2023faithfulness,lanham2023measuring,lyu2023faithful} have explored defining faithfulness for natural language explanations. i) \textit{Faithfulness of token importance explanations:} We use the counterfactual test~\citep{atanasova2023faithfulness} for NLEs by intervening on input tokens and checking whether the explanation reflects these tokens. Specifically, we replace identified importance tokens in the explanation with synonyms and check whether the new explanation reflects these changes. Faithfulness is then quantified by the rank agreement (Eq.~\ref{equation:tokenrankagreement}) between the new explanations and the expected explanation with intervened tokens. ii) \textit{Faithfulness of chain of thought explanations:} Recent works that explored the topic of faithfulness in CoT explanations don't explicitly quantify the faithfulness of an individual explanation. Hence, we follow suit and follow \citet{lanham2023measuring} to measure faithfulness at a dataset level. In our experiments, we use a strategy called ``Early Answering'' proposed by \citet{lanham2023measuring} to measure the faithfulness of CoT explanations. It involves truncating the previously collected reasoning samples and prompting the model to answer the question with the partial CoT rather than the complete one, \ie for a question $Q$ and CoT $[s_1, s_2, \dots s_n]$, the model is prompted to answer with $Q + s_1$, $Q + s_1 + s_2$, until, $Q + s_1 + s_2 \dots + s_n$. After collecting answers with each truncation of the CoT, we measure how often the model comes to the same conclusion as it did with the complete CoT. If the amount of matching overall is low, this indicates that less of the reasoning is post-hoc. If the reasoning is not post-hoc, there are fewer ways for it to be unfaithful than there are for reasoning which is post-hoc~\citep{lanham2023measuring}.

\xhdr{Implementation details}~To run the paraphrase probing uncertainty, we formulate 10 semantically equivalent paraphrases of every question to measure uncertainty using sample probing. In the model probing uncertainty experiment, we sample five natural language explanations at a temperature of $1.0$. To compute the rank agreement of token importance explanations, we use the top-3 words \ie $k = 3$. We run on a randomly sampled subset of 100 samples for each dataset. See the Appendix for more implementation details.


\subsection{Results}
Next, we discuss experimental results to answer questions (RQ1-RQ4) about uncertainty in explanations.
\begin{figure}[t]
  \centering
  \begin{subfigure}{0.45\textwidth}
    \includegraphics[width=\linewidth]{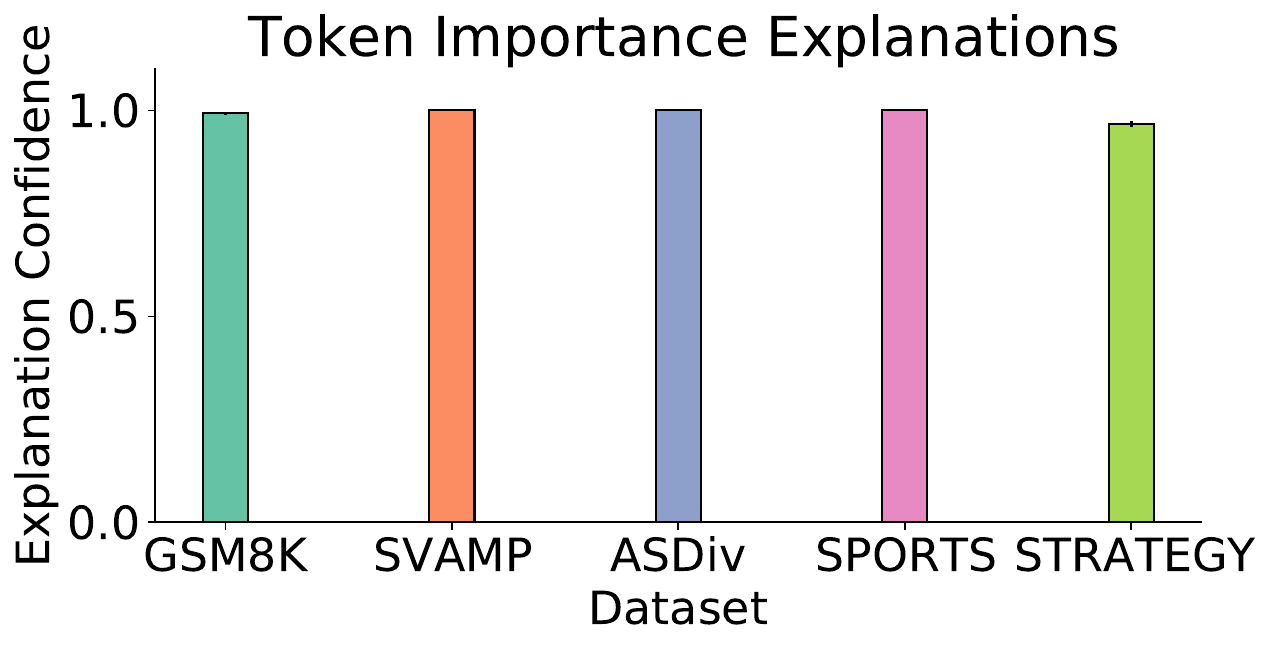}
  \end{subfigure}\hfill
  \begin{subfigure}{0.45\textwidth}
    \includegraphics[width=\linewidth]{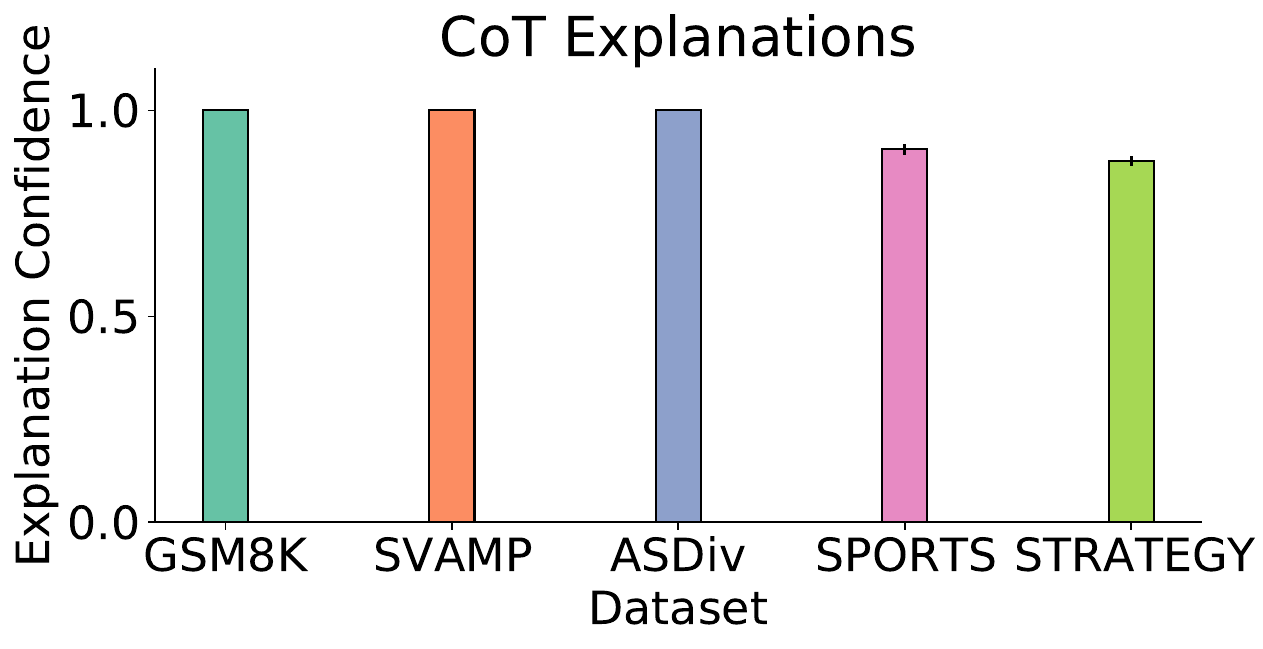}
  \end{subfigure}
  \caption{\small Verbalized explanation confidence of Token Importance and CoT explanations on three math word problems and two commonsense reasoning datasets. We observe that the verbalized explanation confidence is mostly high for explanations across all five datasets.}
  \label{fig:verbalizedoverconfidence}
\end{figure}

\looseness=-1\xhdr{RQ1) Analyzing verbalized uncertainty} Verbalized confidence scores of both natural language explanation methods are almost always 100\%. It raises questions about whether these uncertainty estimates are reliable. If the confidence in every explanation is the same, it is impossible to know when to trust the generated explanation and when not to. Our results in Figure 4 show that, on average, across both explanation methods and five datasets, the verbalized confidence is 94.46\%. Our analysis of these methods uncovers that LLMs often exhibit a high degree of overconfidence when verbalizing their uncertainty in explanations. 
The verbalized uncertainty for commonsense reasoning datasets is lower than math word problem datasets but still very close to 100\% with little standard deviation.
\begin{figure*}
  \centering
  \begin{subfigure}{0.49\textwidth}
    \includegraphics[width=\linewidth]{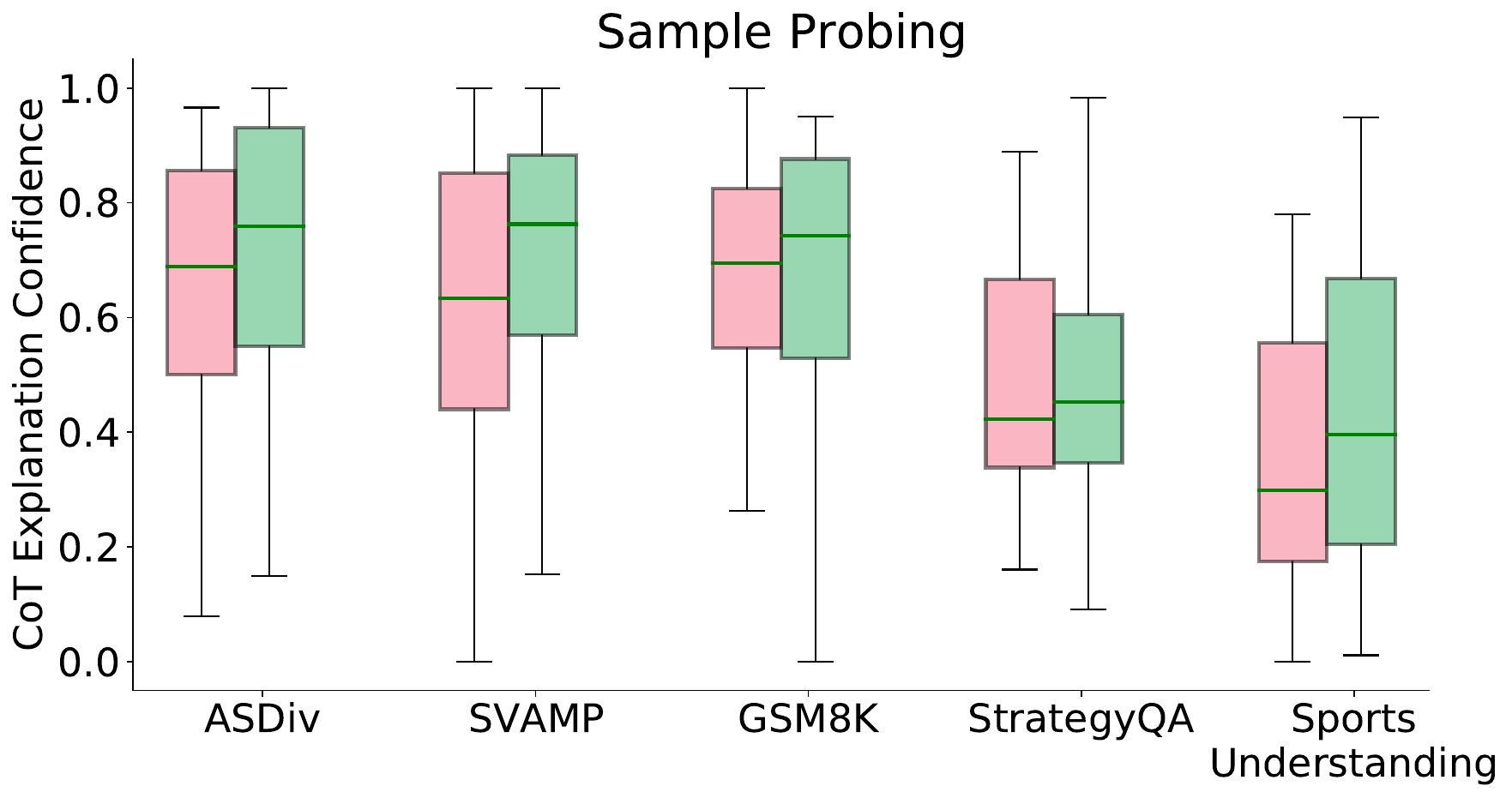}
  \end{subfigure}  
  \begin{subfigure}{0.49\textwidth}
    \includegraphics[width=\linewidth]{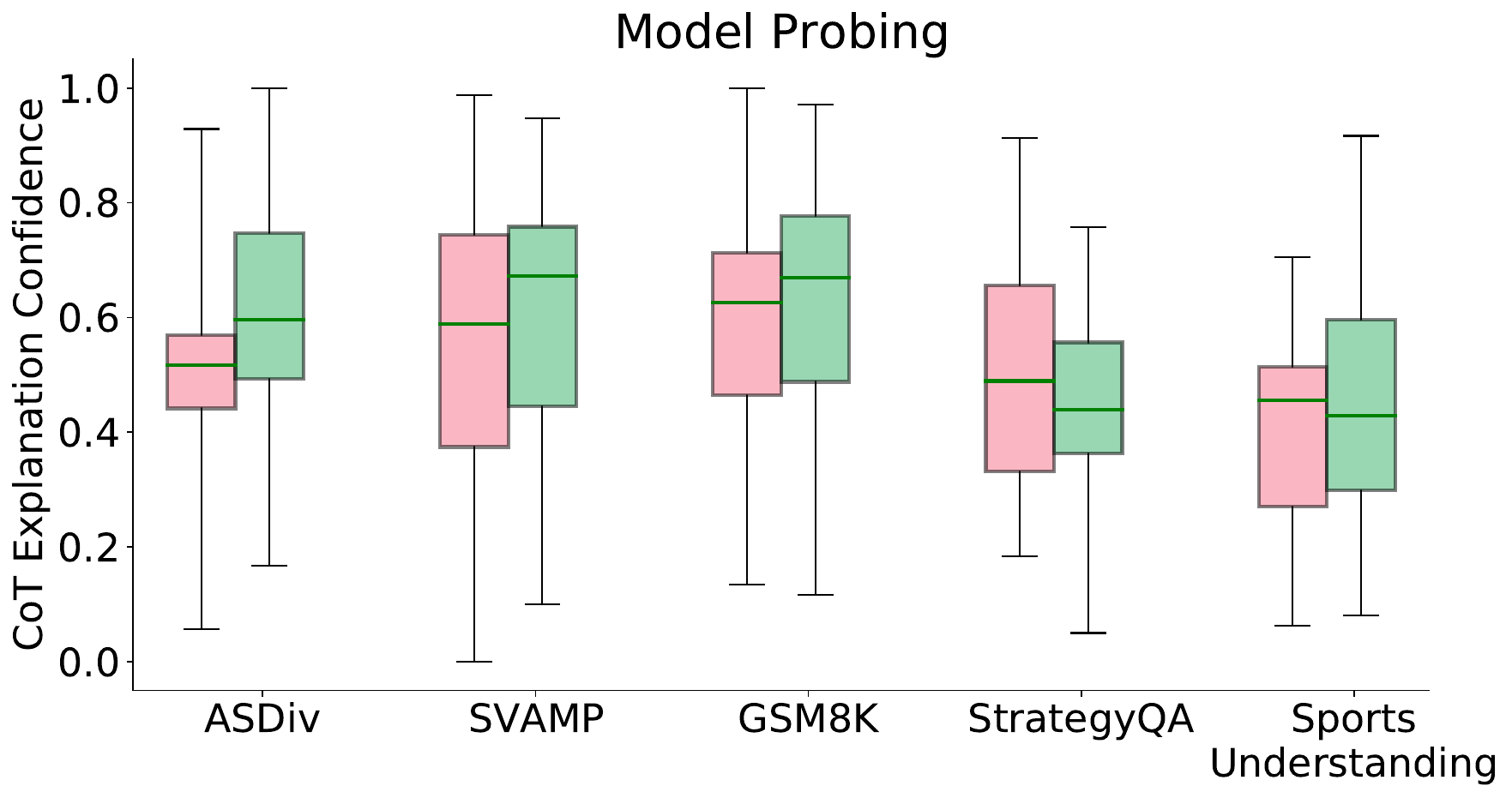}
  \end{subfigure}
  \hfill
  \begin{subfigure}{0.36\textwidth}
    \includegraphics[width=\linewidth]{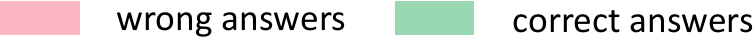}
  \end{subfigure}
  \caption{\small Chain of thought explanation confidence distributions on three math word problems and two commonsense reasoning datasets using \textsc{Gpt}-3.5. On average, across two probing strategies and five datasets, correct answers (in green) obtain higher explanation confidence than wrong answers (in red). See Table~\ref{tab:ttest} in appendix for t-test statistics comparing explanation confidence scores of correct and incorrect answers to different datasets.}
  \vspace{-0.1in}
  \label{fig:cotaccuracy}
\end{figure*}

\begin{figure}[H]
  \centering
  \begin{minipage}{0.42\textwidth}
    \centering
    \includegraphics[width=\textwidth]{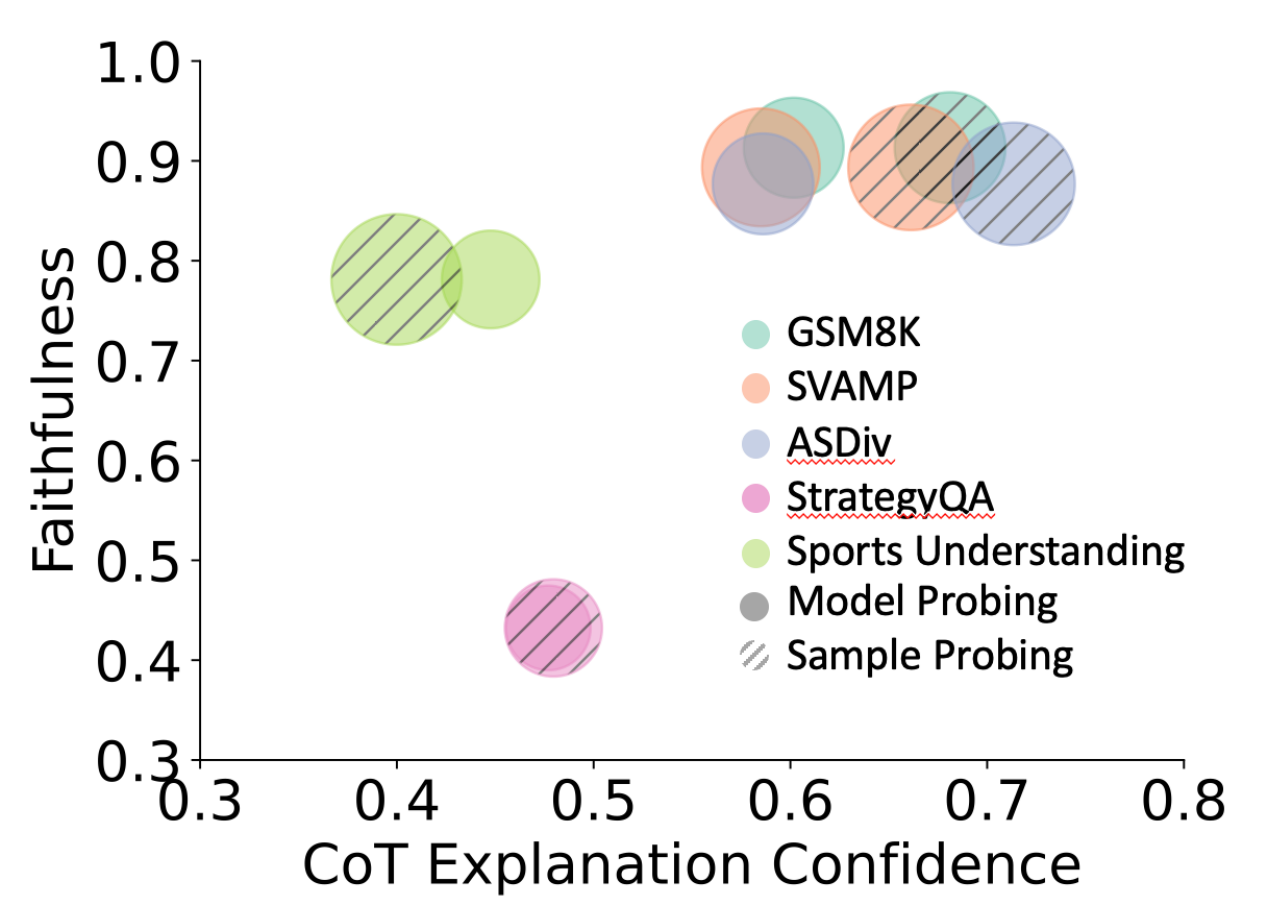}
    \caption{\small Mean explanation confidence for CoT explanations generated using Instruct\textsc{Gpt} for five datasets. We find that the explanation confidence is positively correlated with faithfulness for four datasets, \ie highly confident explanations tend to be more faithful. The circle size denotes the deviation in the confidence.}
    \label{fig:cotfaithfulness}
  \end{minipage}\hfill
  \begin{minipage}{0.55\textwidth}
    \centering
    \includegraphics[width=\textwidth]{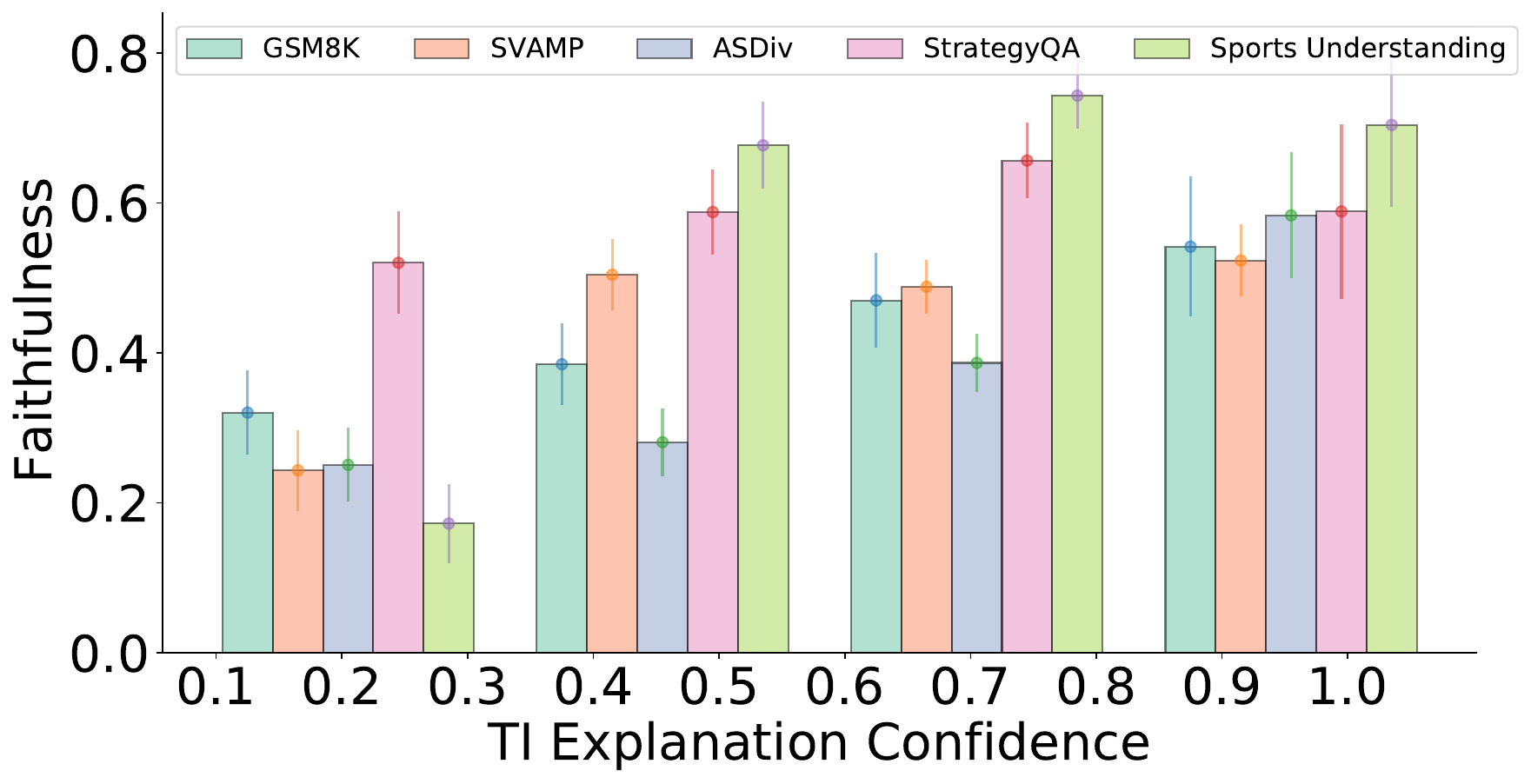}
    \caption{\small Mean explanation confidence for token importance explanations generated using Instruct\textsc{Gpt} for five datasets. We find that the explanation confidence is positively correlated with faithfulness, \ie highly confident explanations tend to be more faithful.}
    \label{fig:fifaithfulness}
  \end{minipage}
\end{figure}

\looseness=-1\xhdr{RQ2) Less uncertain explanations are more faithful} A model's explanation is said to be faithful if it reflects the true reasoning behind the prediction. For token importance explanations, we swap important words in explanations with synonyms and check if the corresponding replacements are reflected in the new explanation. In Fig.~\ref{fig:fifaithfulness}, we demonstrate that explanation confidence is correlated with faithfulness, and highly confident (certain) explanations are more faithful. In addition, we find a similar trend between the CoT explanation confidence and its faithfulness (see Fig.~\ref{fig:cotfaithfulness}) and find that increased mean explanation confidence lead to an increase in the faithfulness of an explanation for most datasets. Our observations suggest that uncertainty estimation can be used as a test for the faithfulness of NLE, \ie whether the explanation reflects the true reasoning process of the model.

\looseness=-1\xhdr{RQ3) Correct answers have more certain explanations} Across five datasets and two probing uncertainty metrics, Fig.~\ref{fig:cotaccuracy} shows that explanations of correct answers have higher explanation confidence compared to explanations of wrong answers. Our observation aligns with the general expectation that models tend to provide more reliable and confident explanations when they make correct predictions as opposed to incorrect ones. 

\begin{figure}[h]
  \centering
  \begin{minipage}{0.45\textwidth}
    \centering
    \includegraphics[width=\textwidth]{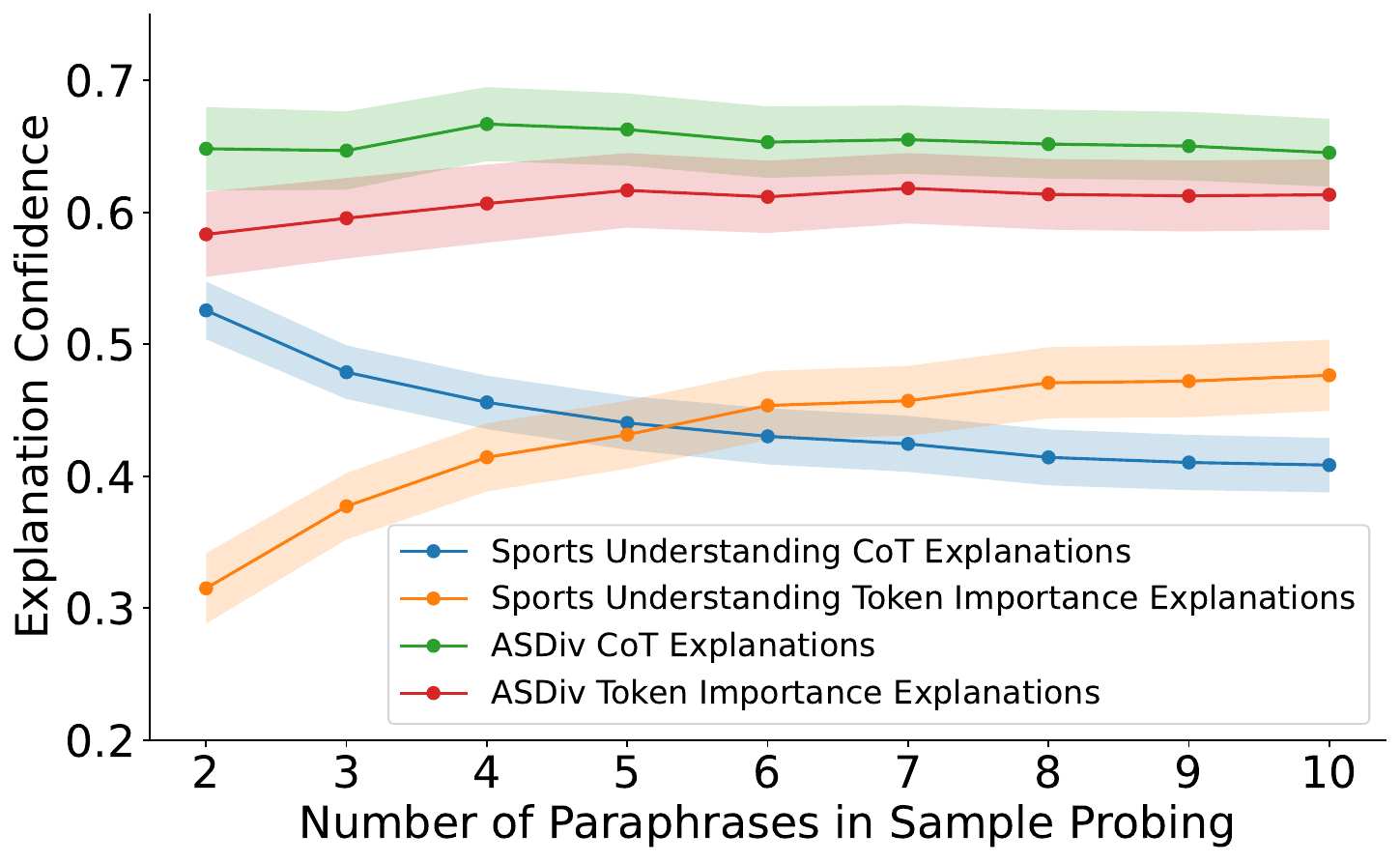}
    \caption{\small The effect of the number of paraphrased samples of the original $Q$ on the mean explanation confidence of CoT and TI explanations generated from Instruct\textsc{Gpt} for Sports Understanding and ASDiv datasets. We observe that the confidence saturates as we increase the number of paraphrased samples.}
    \label{fig:sampleprobingablation}
  \end{minipage}\hfill
  \begin{minipage}{0.49\textwidth}
    \centering
    \includegraphics[width=\textwidth]{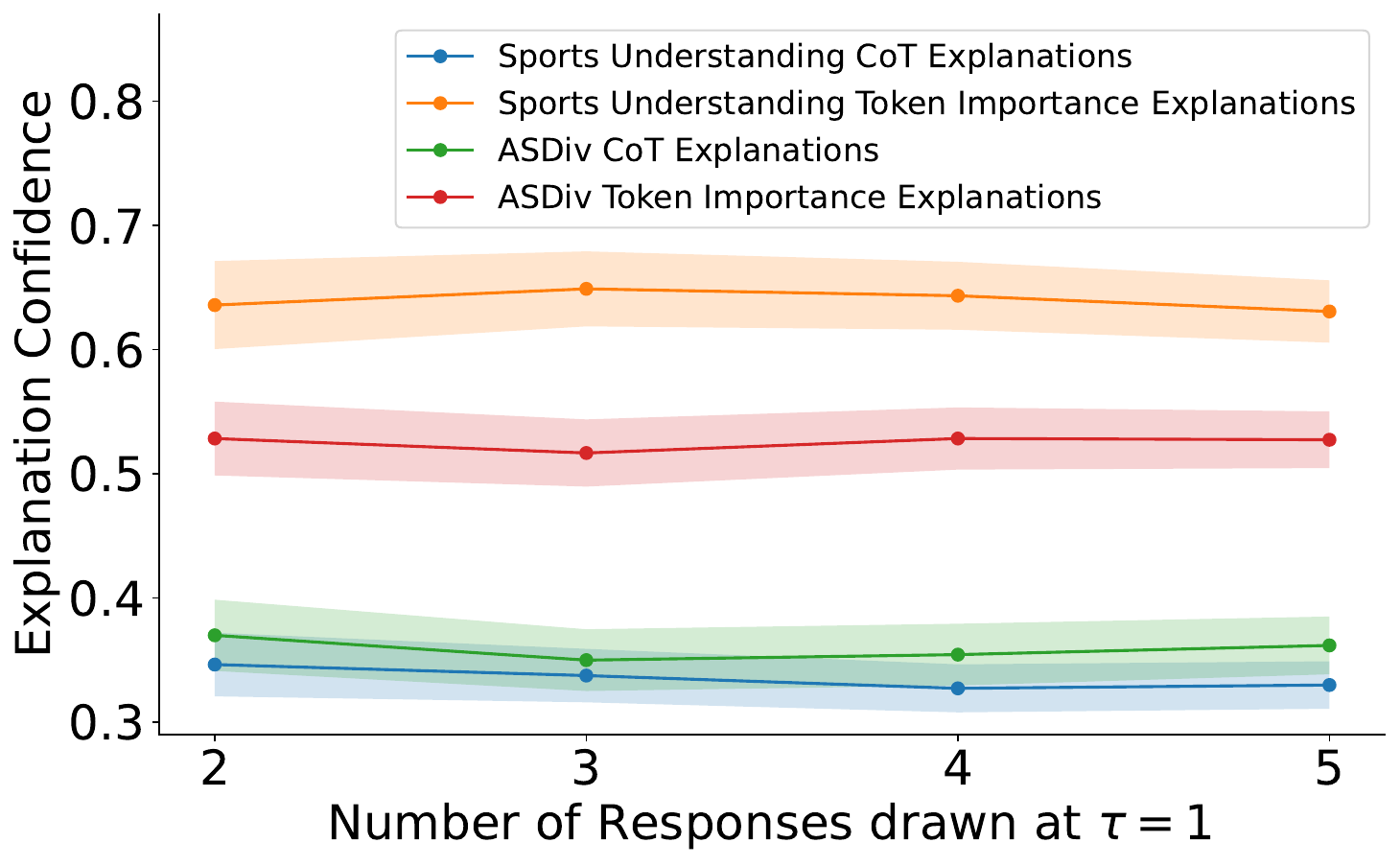}
    \caption{\small The effect of the number of responses drawn at $\tau=1$ on the mean explanation confidence of CoT and TI explanations generated from Instruct\textsc{Gpt} for Sports Understanding and ASDiv datasets. We observe that the confidence remains consistent irrespective of the number of responses generated using Instruct\textsc{Gpt}.}
    \label{fig:modelprobingablation}
  \end{minipage}
\end{figure}

\xhdr{RQ4) Ablation study} We conduct ablation on three key components of our proposed probing metrics i) the number of paraphrases we generate in sample probing, ii) the number of samples we generate at temperature $\tau=1$ in model probing, and iii) different LLMs (see Figs.~\ref{fig:modelcomparision-sampleprobing}-\ref{fig:modelcomparision-modelprobing} in appendix for more details and results). Results in Figure~\ref{fig:sampleprobingablation} show that the explanation confidence saturates as we increase the number of paraphrases of the original question $Q$ and our chosen value of 10 is well justified. In addition, we observe that the explanation confidence using our proposed model probing technique shows similar behavior irrespective of the number of responses we generate using the LLM at $\tau=1$ (Figure~\ref{fig:modelprobingablation}). These findings explain our choice of hyperparameters in quantifying the uncertainty in explanations generated using different NLE techniques.

\section{Conclusions}
\vspace{-0.06in}
\looseness=-1 While improving the explainability of LLMs is crucial to establishing user trust, and better understanding the limitations and unintended biases present in LLMs, it is crucial to quantify the reliability of the generated explanations using uncertainty estimates. In this work, we present a novel way to estimate the uncertainty of natural language explanations (NLEs) using verbalized and probing techniques. Specifically, we propose uncertainty metrics to quantify the confidence of generated NLEs from LLMs and compare their reliability. We test the effectiveness of our metrics on math word problem and commonsense reasoning datasets and find that i) LLMs exhibit a high degree of overconfidence when verbalizing their uncertainty in explanations, ii) explanation confidence is positively correlated with explanation faithfulness, and iii) correct predictions tend to have more certain CoT explanations compared to incorrect predictions. Our work paves the way for several exciting future works in understanding the uncertainty of the natural language explanations generated by LLMs.

\bibliographystyle{apalike}
\bibliography{aistats2024/references}

\newpage
\appendix
\section{Appendix}
\begin{figure}[h]
  \centering
  \begin{subfigure}{0.42\textwidth}
    \includegraphics[width=\linewidth]{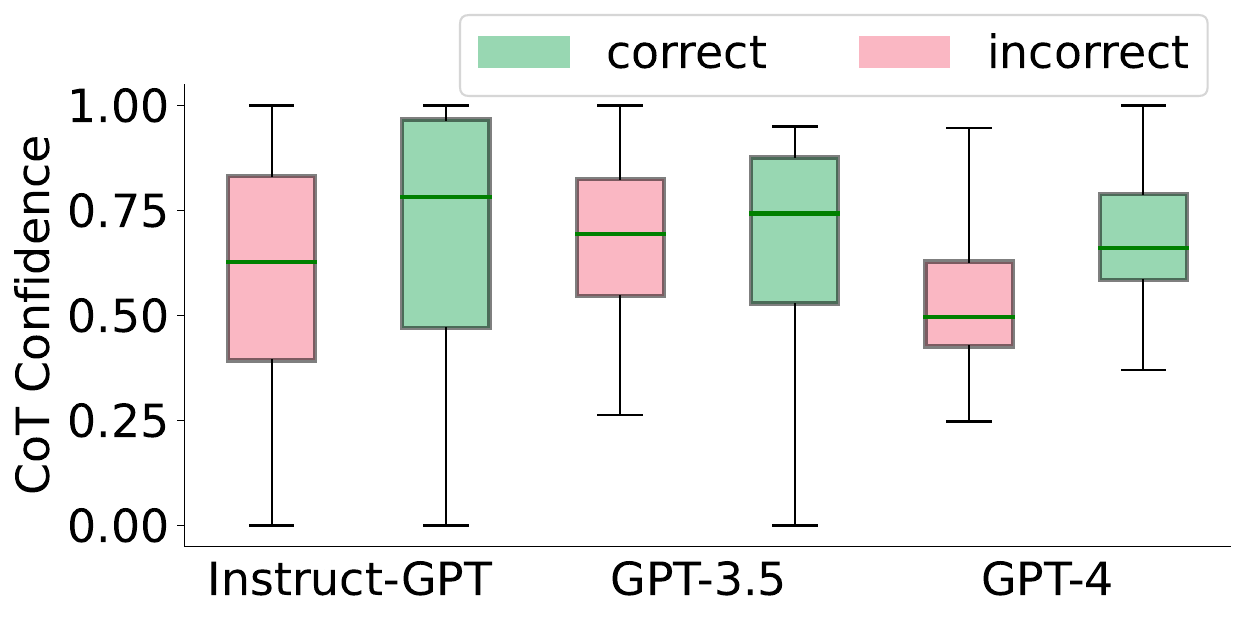}
  \end{subfigure}\hfill
  \caption{Comparison of chain of thought explanation uncertainty using sample probing across Instruct\textsc{Gpt}, \textsc{Gpt}-3.5, and \textsc{Gpt}-4 models on GSM8K dataset. We observe that the trend of correct answers having less uncertain explanations holds true across models.}
  \label{fig:modelcomparision-sampleprobing}
\end{figure}

\begin{figure}[h]
  \centering
  \begin{subfigure}{0.42\textwidth}
    \includegraphics[width=\linewidth]{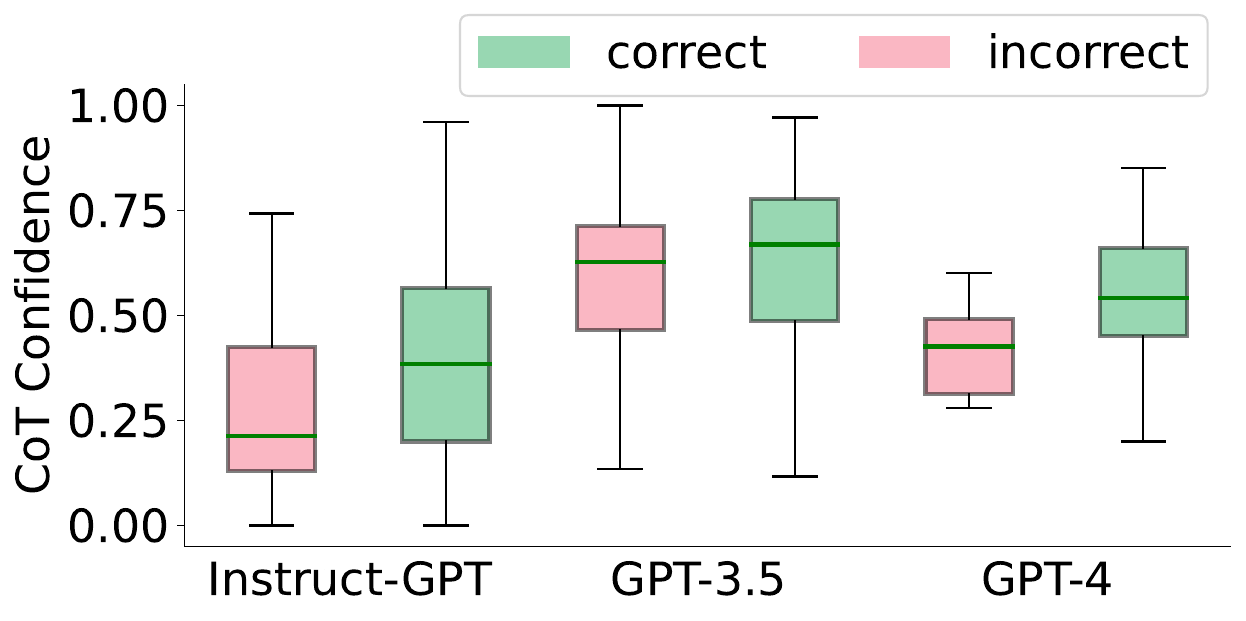}
  \end{subfigure}\hfill
  \caption{Comparision of chain of thought explanation uncertainty using model probing across Instruct\textsc{Gpt}, \textsc{Gpt}-3.5, and \textsc{Gpt}-4 models on GSM8K dataset. We observe that the trend of correct answers having less uncertain explanations holds true across models.}
  \label{fig:modelcomparision-modelprobing}
\end{figure}

\begin{figure*}[h]
    \begin{dialogbox}
Read the question, and output the words important for your final answer, sorted in descending order of importance. The output format is as follows:\vspace{0.05in}

\textbf{1. [Word 1 here]} \\
\textbf{2. [Word 2 here]} \\
\dots \\
\dots \\
\textbf{N. [Word N here]} \\
\textcolor{purple}{\textbf{Final Answer and Overall Confidence (0-100):}} [Your answer as a number here], [Your confidence here]\%. Provide the answer in aforementioned format, and nothing else.
    \end{dialogbox}
    \caption{The prompt $Q_e$ prepended to the question $Q$ to elicit a token importance explanation $TI$ along with an answer $A$.}\label{fig:featureimportanceexplanation}
\end{figure*}

\begin{figure*}[h]
\begin{dialogbox}
Read the question, give your answer by analyzing step by step, and assign a confidence level to each step and the final answer. The output format is as follows:\\
\textbf{Step 1}: [Your reasoning here] \\
\textbf{Step 2}: [Your reasoning here] \\
\textbf{Step 3}: \\ ...\\
...\\
\textbf{Step N}: [Your reasoning here] \\
\textcolor{purple}{\textbf{Final Answer and Overall Confidence (0-100):}} [Your answer as a number here], [Your confidence here]\%
Note: The confidence indicates the degree of certainty you have about your reasoning. For instance, if your confidence level is 80\%, it means you are 80\% certain that your reasoning is correct.
Provide the answer in aforementioned format, and nothing else.
\end{dialogbox}
    \caption{The prompt $Q_e$ prepended to the question $Q$ to elicit a chain of thought explanation $CoT$ along with an answer $A$.}\label{fig:cotexplanation}
\end{figure*}

\begin{table*}[ht]
\centering
\caption{T-Test Result Comparing Explanation Confidence Scores of Correct and Incorrect Answers using \textsc{Gpt}-3.5 and Instruct\textsc{Gpt} models for Chain of Thought Explanations of GSM8K dataset.}

\begin{tabular}{|c|c|c|c|}
\hline
\textbf{Dataset}& \textbf{Uncertainty Metric} & \textbf{T-Statistic} & \textbf{P-Value} \\
\hline
\multirow{2}{*}{GSM8K} &  Sample Probing & -0.0977 & 0.9224 \\
& Model Probing & 0.7400 & 0.4611 \\
\hline
\multirow{2}{*}{SVAMP} &  Sample Probing & 1.7913 & 0.0763 \\
& Model Probing & 1.2307 & 0.2214 \\
\hline
\multirow{2}{*}{ASDiv} &  Sample Probing & 1.3031 & 0.1959 \\
& Model Probing & 1.7922 & 0.0765 \\
\hline
\multirow{2}{*}{StrategyQA} &  Sample Probing & -0.2752 & 0.7838 \\
& Model Probing & -0.9779 & 0.3305 \\
\hline
\multirow{2}{*}{Sports Understanding} &  Sample Probing & 1.3941 & 0.1665 \\
& Model Probing & 1.0851 & 0.2806 \\
\hline
\end{tabular}
\\
\vspace{5pt}
\text{(i) \textsc{Gpt}-3.5}
\vspace{5pt}
\\
\begin{tabular}{|c|c|c|c|}
\hline
\textbf{Dataset}& \textbf{Uncertainty Metric} & \textbf{T-Statistic} & \textbf{P-Value} \\
\hline
\multirow{2}{*}{GSM8K} &  Sample Probing & 1.5694 & 0.1198 \\
& Model Probing & 3.2404 & 0.0016 \\
\hline
\multirow{2}{*}{SVAMP} &  Sample Probing & 2.6388 & 0.0097 \\
& Model Probing & 0.7660 & 0.4455 \\
\hline
\multirow{2}{*}{ASDiv} &  Sample Probing & 3.7558 & 0.0003 \\
& Model Probing & 5.1783 & 0.0000 \\
\hline
\multirow{2}{*}{StrategyQA} &  Sample Probing & -0.1642 & 0.8699 \\
& Model Probing & -0.1015 & 0.9194 \\
\hline
\multirow{2}{*}{Sports Understanding} &  Sample Probing & -0.8499 & 0.3975 \\
& Model Probing & 0.6971 & 0.4874 \\
\hline
\end{tabular}
\\
\vspace{5pt}
\text{(ii) Instruct\textsc{Gpt}}
\vspace{5pt}
\label{tab:ttest}
\end{table*}

\subsection*{Prompts}

The questions used to generate chain of thought and token importance explanations are described in \ref{fig:cotexplanation} and \ref{fig:featureimportanceexplanation} respectively. For sample probing and model probing uncertainty, we further tailor the prompt according to the dataset. Tailoring the question prompt helps in parsing answers and explanations from generated text. The prompts used are as follows GSM8K \ref{fig:gsm8kcotexplanation} \ref{fig:gsm8kfeatureimportanceexplanation}, ASDiv \ref{fig:asdivcotexplanation} \ref{fig:asdivfeatureimportanceexplanation}, SVAMP \ref{fig:svampcotexplanation} \ref{fig:svampfeatureimportanceexplanation}, StrategyQA \ref{fig:strategyqacotexplanation} \ref{fig:strategyqafeatureimportanceexplanation}, and Sports Understanding \ref{fig:sportscotexplanation} \ref{fig:sportsfeatureimportanceexplanation}.

\begin{figure*}[h]
    \begin{dialogbox}
Read the question, and output the words important for your final answer, sorted in descending order of importance. The output format is as follows:\vspace{0.05in}

\textbf{1. [Word 1 here]} \\
\textbf{2. [Word 2 here]} \\
\dots \\
\dots \\
\textbf{N. [Word N here]} \\
\textcolor{purple}{\textbf{Final Answer and Overall Confidence (0-100):}} [Your answer as a number here], [Your confidence here]\%. Provide the answer in aforementioned format, and nothing else.
    \end{dialogbox}
    \caption{\textbf{GSM8K} dataset. The prompt $Q_e$ prepended to the paraphrased question $Q$ to generate a token importance explanation $TI$ along with an answer $A$ in sample probing and model probing uncertainty experiments.}\label{fig:gsm8kfeatureimportanceexplanation}
\end{figure*}

\begin{figure*}[h]
\begin{dialogbox}
Read the question, give your answer by analyzing step by step, and assign a confidence level to each step and the final answer. The output format is as follows:\\
\textbf{Step 1}: [Your reasoning here], Confidence: [Your confidence here]\% \\
\textbf{Step 2}: [Your reasoning here], Confidence: [Your confidence here]\% \\
\textbf{Step 3}: \\ ...\\
...\\
\textbf{Step N}: [Your reasoning here], Confidence: [Your confidence here]\% \\
\textcolor{purple}{\textbf{Final Answer and Overall Confidence (0-100):}} [Your answer as a number here], [Your confidence here]\%
Note: The confidence indicates the degree of certainty you have about your answer. For instance, if your confidence level is 80\%, it means you are 80\% certain that your answer is correct.
Provide the answer in aforementioned format, and nothing else.
\end{dialogbox}
    \caption{\textbf{GSM8K} dataset. The prompt $Q_e$ prepended to the paraphrased question $Q$ to elicit a chain of thought explanation $CoT$ along with an answer $A$ in sample probing and model probing uncertainty experiments.}\label{fig:gsm8kcotexplanation}
\end{figure*}

\begin{figure*}[h]
    \begin{dialogbox}
Read the question, and output the words important for your final answer, sorted in descending order of importance. The output format is as follows:\vspace{0.05in}

\textbf{1. [Word 1 here]} \\
\textbf{2. [Word 2 here]} \\
\dots \\
\dots \\
\textbf{N. [Word N here]} \\
\textcolor{purple}{\textbf{Final Answer and Overall Confidence (0-100):}} [Your answer as a number here], [Your confidence here]\%. Provide the answer in aforementioned format, and nothing else.
    \end{dialogbox}
    \caption{\textbf{ASDiv} dataset. The prompt $Q_e$ prepended to the paraphrased question $Q$ to generate a token importance explanation $TI$ along with an answer $A$ in sample probing and model probing uncertainty experiments.}\label{fig:asdivfeatureimportanceexplanation}
\end{figure*}

\begin{figure*}[h]
\begin{dialogbox}
Read the question, give your answer by analyzing step by step, and assign a confidence level to each step and the final answer. The output format is as follows:\\
\textbf{Step 1}: [Your reasoning here], Confidence: [Your confidence here]\% \\
\textbf{Step 2}: \\ ...\\
\textbf{Step 3}: \\ ...\\
...\\
\textbf{Step N}: \\ ...\\
\textcolor{purple}{\textbf{Final Answer and Overall Confidence (0-100):}} [Your answer as a number here], [Your confidence here]\%
Note: The confidence indicates the degree of certainty you have about your answer. For instance, if your confidence level is 80\%, it means you are 80\% certain that your answer is correct.
Provide the answer in aforementioned format, and nothing else.
\end{dialogbox}
    \caption{\textbf{ASDiv} dataset. The prompt $Q_e$ prepended to the paraphrased question $Q$ to elicit a chain of thought explanation $CoT$ along with an answer $A$ in sample probing and model probing uncertainty experiments.}\label{fig:asdivcotexplanation}
\end{figure*}

\begin{figure*}[h]
    \begin{dialogbox}
Read the question, and output the words important for your final answer, sorted in descending order of importance. The output format is as follows:\vspace{0.05in}

\textbf{1. [Word 1 here]} \\
\textbf{2. [Word 2 here]} \\
\dots \\
\dots \\
\textbf{N. [Word N here]} \\
\textcolor{purple}{\textbf{Final Answer and Overall Confidence (0-100):}} [Your answer as a number here], [Your confidence here]\%. Provide the answer in aforementioned format, and nothing else.
    \end{dialogbox}
    \caption{\textbf{SVAMP} dataset. The prompt $Q_e$ prepended to the paraphrased question $Q$ to generate a token importance explanation $TI$ along with an answer $A$ in sample probing and model probing uncertainty experiments.}\label{fig:svampfeatureimportanceexplanation}
\end{figure*}

\begin{figure*}[h]
\begin{dialogbox}
Read the question, give your answer by analyzing step by step, and assign a confidence level to each step and the final answer. The output format is as follows:\\
\textbf{Step 1}: [Your reasoning here], Confidence: [Your confidence here]\% \\
\textbf{Step 2}: \\ ...\\
\textbf{Step 3}: \\ ...\\
...\\
\textbf{Step N}: \\ ...\\
\textcolor{purple}{\textbf{Final Answer and Overall Confidence (0-100):}} [Your answer as a number here], [Your confidence here]\%
Note: The confidence indicates the degree of certainty you have about your answer. For instance, if your confidence level is 80\%, it means you are 80\% certain that your answer is correct.
Provide the answer in aforementioned format, and nothing else.
\end{dialogbox}
    \caption{\textbf{SVAMP} dataset. The prompt $Q_e$ prepended to the paraphrased question $Q$ to elicit a chain of thought explanation $CoT$ along with an answer $A$ in sample probing and model probing uncertainty experiments.}\label{fig:svampcotexplanation}
\end{figure*}

\begin{figure*}[h]
    \begin{dialogbox}
Read the question, and output the words important for your final answer, sorted in descending order of importance. The output format is as follows:\vspace{0.05in}

\textbf{1. [Word 1 here]} \\
\textbf{2. [Word 2 here]} \\
\dots \\
\dots \\
\textbf{N. [Word N here]} \\
\textcolor{purple}{\textbf{Final Answer and Overall Confidence (0-100):}} [Your answer Yes/No here], [Your confidence here]\%. Provide the answer in aforementioned format, and nothing else.
    \end{dialogbox}
    \caption{\textbf{StrategyQA} dataset. The prompt $Q_e$ prepended to the paraphrased question $Q$ to generate a token importance explanation $TI$ along with an answer $A$ in sample probing and model probing uncertainty experiments.}\label{fig:strategyqafeatureimportanceexplanation}
\end{figure*}

\begin{figure*}[h]
\begin{dialogbox}
Read the question, give your answer by analyzing step by step, and assign a confidence level to each step and the final answer. The output format is as follows:\\
\textbf{Step 1}: [Your reasoning here], Confidence: [Your confidence here]\% \\
\textbf{Step 2}: \\ ...\\
\textbf{Step 3}: \\ ...\\
...\\
\textbf{Step N}: \\ ...\\
\textcolor{purple}{\textbf{Final Answer and Overall Confidence (0-100):}} [Your answer Yes/No here], [Your confidence here]\%
Note: The confidence indicates the degree of certainty you have about your answer. For instance, if your confidence level is 80\%, it means you are 80\% certain that your answer is correct.
Provide the answer in aforementioned format, and nothing else.
\end{dialogbox}
    \caption{\textbf{StrategyQA} dataset. The prompt $Q_e$ prepended to the paraphrased question $Q$ to elicit a chain of thought explanation $CoT$ along with an answer $A$ in sample probing and model probing uncertainty experiments.}\label{fig:strategyqacotexplanation}
\end{figure*}

\begin{figure*}[h]
    \begin{dialogbox}
Read the question, and output the words important for your final answer, sorted in descending order of importance. The output format is as follows:\vspace{0.05in}

\textbf{1. [Word 1 here]} \\
\textbf{2. [Word 2 here]} \\
\dots \\
\dots \\
\textbf{N. [Word N here]} \\
\textcolor{purple}{\textbf{Final Answer and Overall Confidence (0-100):}} [Your answer plausible / implausible here], [Your confidence here]\%. Provide the answer in aforementioned format, and nothing else.
    \end{dialogbox}
    \caption{\textbf{Sports Understanding} dataset. The prompt $Q_e$ prepended to the paraphrased question $Q$ to generate a token importance explanation $TI$ along with an answer $A$ in sample probing and model probing uncertainty experiments.}\label{fig:sportsfeatureimportanceexplanation}
\end{figure*}

\begin{figure*}[h]
\begin{dialogbox}
Read the question, give your answer by analyzing step by step, and assign a confidence level to each step and the final answer. The output format is as follows:\\
\textbf{Step 1}: [Your reasoning here], Confidence: [Your confidence here]\% \\
\textbf{Step 2}: ...\\
\textbf{Step 3}: ...\\
...\\
\textbf{Step N}: ...\\
\textcolor{purple}{\textbf{Final Answer and Overall Confidence (0-100):}} [Your answer plausible / implausible here], [Your confidence here]\%
Note: The confidence indicates the degree of certainty you have about your answer. For instance, if your confidence level is 80\%, it means you are 80\% certain that your answer is correct.
Provide the answer in aforementioned format, and nothing else.
\end{dialogbox}
    \caption{\textbf{Sports Understanding} dataset. The prompt $Q_e$ prepended to the paraphrased question $Q$ to elicit a chain of thought explanation $CoT$ along with an answer $A$ in sample probing and model probing uncertainty experiments.}\label{fig:sportscotexplanation}
\end{figure*}

\subsection*{Paraphrased Questions in Sample Probing}

Semantically equivalent paraphrased questions are generated using \textsc{InstructGpt} using the following prompt - \textit{"Paraphrase the question into 10 different forms with the same meaning, and share them as a Python list of double quotes enclosed strings"}. An example is shown in \ref{tab:sampleprobingexample}.

\begin{table*}
    \centering
    \caption{Paraphrased Samples for a question in GSM8K math word problem dataset. The original question is \textit{"How many signatures do the sisters need to collect to reach their goal?"}}
    \begin{tabular}{|p{0.8\linewidth}|}
        \hline
        What is the number of signatures the sisters need to collect to reach their goal? \\
        \hline
        How many signatures must the sisters acquire to reach their goal? \\
        \hline
        What is the amount of signatures the sisters need to collect to reach their goal? \\
        \hline
        How many signatures do the sisters have to collect to reach their goal? \\
        \hline
        What is the total number of signatures the sisters need to collect to reach their goal? \\
        \hline
        How many signatures do the sisters require to reach their goal? \\
        \hline
        What is the quantity of signatures the sisters need to collect to reach their goal? \\
        \hline
        How many signatures do the sisters need to gather to reach their goal? \\
        \hline
        What is the sum of signatures the sisters need to collect to reach their goal? \\
        \hline
        How many signatures do the sisters need to acquire to reach their goal? \\
        \hline
    \end{tabular}
    \label{tab:sampleprobingexample}
\end{table*}

\end{document}